\documentclass[10pt,twocolumn]{article}
\usepackage{graphicx}
\usepackage{amsmath,amssymb,times,cite}

\newcommand{\op}[1]{\operatorname{#1}}
\newcommand{\bg}[1]{\boldsymbol{#1}} 

\newcommand{\bm}[1]{\mathbf{#1}} 
\newcommand\T{{\mathpalette\raiseT\intercal}}
\newcommand\raiseT[2]{%
\setbox0\hbox{$#1{#2}$}\raise\dp0\box0}

\usepackage{multirow}
\usepackage{booktabs} 
\usepackage{xcolor}

\usepackage{amsthm}

\newtheorem{lemma}{Lemma}

\newtheorem{proposition}{Proposition}

\title{\Large\textbf{Fixed-Point Graph Convolutional Networks Against Adversarial Attacks}}
\author{Shakib Khan, A. Ben Hamza, and Amr Youssef\\
Concordia Institute for Information Systems Engineering\\
Concordia University, Montreal}

\date{}

\begin{document}
\maketitle

\begin{abstract}
Adversarial attacks present a significant risk to the integrity and performance of graph neural networks, particularly in tasks where graph structure and node features are vulnerable to manipulation. In this paper, we present a novel model, called fixed-point iterative graph convolutional network (Fix-GCN), which achieves robustness against adversarial perturbations by effectively capturing higher-order node neighborhood information in the graph without additional memory or computational complexity. Specifically, we introduce a versatile spectral modulation filter and derive the feature propagation rule of our model using fixed-point iteration. Unlike traditional defense mechanisms that rely on additional design elements to counteract attacks, the proposed graph filter provides a flexible-pass filtering approach, allowing it to selectively attenuate high-frequency components while preserving low-frequency structural information in the graph signal. By iteratively updating node representations, our model offers a flexible and efficient framework for preserving essential graph information while mitigating the impact of adversarial manipulation. We demonstrate the effectiveness of the proposed model through extensive experiments on various benchmark graph datasets, showcasing its resilience against adversarial attacks.
\end{abstract}

\bigskip
\noindent\textbf{Keywords}: Graph neural networks; adversarial attacks; higher-order convolution; fixed-point iteration.

\section{Introduction}
Graph neural networks (GNNs), particularly graph convolutional networks (GCNs)~\cite{Kipf2017GCN} and their variants~\cite{Wu2019SGC,Abu2019MixHop,Chen2020GCNII}, have proven effective at learning representations of graph-structured data, demonstrating state-of-the-art performance in a wide variety of real-world applications, including weather forecasting~\cite{Lam2023GraphCast}, biomedicine and healthcare~\cite{Li2022Bio}, traffic forecasting~\cite{zhao2019TGCN}, and cybersecurity~\cite{He2022ILLUMINATI}. The key to the effectiveness of GNNs lies in the neural message passing scheme, which iteratively passes and aggregates feature information from neighboring nodes in the graph. However, despite their effectiveness, GNNs are vulnerable to adversarial attacks~\cite{Zugner2018Nettack,Zugner2019Mettack}, which involve intentionally crafted perturbations to the input graph data, such as modifying the graph structure or altering node features, with the goal of causing the model to make incorrect predictions.

Adversarial attacks on graphs can be classified into poisoning and evasion attacks based on the attacker's objectives and the timing of the attack~\cite{Zugner2018Nettack,Zugner2019Mettack,Jin2021Survey,Sun2022Survey}. Evasion attacks, also known as test-time attacks, aim to deceive the model and compromise its prediction performance at testing time. On the other hand, poisoning attacks, also known as training-time attacks, manipulate the training data to mislead the model during the training phase with the aim of degrading the model performance on downstream tasks. Moreover, poisoning attacks can be broadly categorized into targeted and non-targeted attacks. Targeted attacks, such as Nettack~\cite{Zugner2018Nettack}, aim to misclassify specific nodes in the graph by perturbing the graph structure during the training phase. Rather than focusing on specific nodes, non-targeted attacks, such as Mettack~\cite{Zugner2019Mettack}, aim to degrade the overall performance of GNNs by perturbing the graph structure (i.e., adding or modifying edges) during the training phase. These adversarial attacks pose significant challenges to the integrity and performance of GNN-based systems, especially in critical applications, such as healthcare, cybersecurity and autonomous driving, where trustworthiness and robustness are paramount. Hence, it is crucial to incorporate adversarial defense mechanisms into GNN models and/or develop and design robust and resilient model architectures that can resist adversarial attacks effectively.

While integrating defense mechanisms into GNN models has demonstrated effectiveness in thwarting adversarial attacks~\cite{Wu2019GCNJaccard,Zhu2019RGCN,Entezari2020GCNSVD,Jin2020ProGNN,Alchihabi2023ELRGNN}, these mechanisms often employ additional components that require extensive computations for a successful defense. For instance, GCN-SVD~\cite{Entezari2020GCNSVD} preprocesses the graph adjacency matrix using singular value decomposition (SVD), retaining only the low-frequency components to defend against adversarial attacks on graph structures. The basic idea is to remove high-frequency perturbations that may have been introduced by adversarial attacks while preserving the essential low-frequency structural features. However, since GCN-SVD, tailored specifically for Nettack~\cite{Zugner2018Nettack}, focuses on preprocessing the poisoned graph data, it may not offer consistent efficacy against diverse adversarial attacks. A recent line of work focuses on designing GNN model architectures that are robust against adversarial attacks~\cite{geisler2020reliable,wang2020provably,zhang2020feature,Jin2021SimpGCN,Liu2021ElasticGCN,Chang2021LPF,Lei2022EvenNet,Huang2023MidGCN}. For instance, Mid-GCN~\cite{Huang2023MidGCN} introduces a mid-pass graph filter based on the observation that the eigenvalues of the graph Laplacian within the mid-frequency range are less susceptible to adversarial attacks. However, the mid-frequency information captured by this filter may not sufficiently preserve important structural features of the graph, leading to degraded performance. In addition, the training of Mid-GCN exhibits instability, particularly evident as the spectral radius of its propagation matrix increases with network depth.

In this paper, we introduce a novel and robust model, named fixed-point iterative graph convolutional network (Fix-GCN), designed to withstand diverse adversarial attacks under varying perturbation levels.  The core message-passing mechanism of Fix-GCN is derived by solving a graph filtering system using fixed-point iteration~\cite{Burden2015Numerical}. Our proposed network, belonging to the category of resilient architectures, aims to withstand adversarial perturbations by designing a flexible-pass, higher-order filter that selectively attenuates high-frequency components while preserving low-frequency structural information in the graph signal. This selective attenuation of high-frequency components ensures that the model remains robust and resilient, even when faced with adversarial manipulation. Moreover, by capturing information from higher-order neighbors, Fix-GCN can mitigate the risk posed by direct perturbations on 1-hop neighbors of target nodes, which are often more effective than indirect perturbations (i.e., influencer attacks) on multi-hop neighbors. In addition, similar to the GCN-SVD defense method, our approach with the spectral modulation filter operates on the principle of selectively preserving low-frequency components and discarding high-frequency ones in the graph signal to defend against adversarial attacks while preserving essential graph structural information. Our key contributions can be summarized as follows:
\begin{itemize}
\item We propose a novel spectral modulation filter that provides a mechanism to selectively attenuate high-frequency components while preserving low-frequency structural information in the graph signal.
 \item We present a graph convolutional network architecture with an aggregation mechanism that captures information from higher-order neighbors of graph nodes, while maintaining computational efficiency.
\item Experimental results demonstrate the robustness of the proposed model against adversarial attacks, outperforming competitive baselines across various benchmark datasets.
\end{itemize}

The rest of the paper is organized as follows. In Section~\ref{Related}, we review important related work. In Section~\ref{Method}, we start by presenting the problem statement. Then, we propose a versatile spectral modulation graph filter and introduce a robust graph neural network whose feature propagation rule is derived via fixed-point iteration. In Section~\ref{Experiments}, we present experimental evaluation of the robustness of our model against diverse adversarial attacks in comparison with established baselines. Finally, we conclude in Section~\ref{Conclusion} and point out future work directions.

\section{Related Work} \label{Related}
\noindent\textbf{Adversarial Attacks on GNNs.} The aim of adversarial attacks is to manipulate the behavior of GNN models by introducing small yet impactful changes to the graph structure, node features, or both. These manipulations are often imperceptible, but can have advert effects on the model's predictions and overall performance.  Adversarial attacks can be broadly categorized into targeted and non-targeted attacks, each with distinct objectives. In targeted attacks, such as Nettack~\cite{Zugner2018Nettack}, the adversary's goal is to induce specific misclassifications in the GNN model. Rather than altering the entire graph structure, targeted attacks focus on manipulating the predictions of individual nodes. By perturbing the node features or edges connected to specific target nodes, the attacker aims to deceive the model into producing incorrect predictions for those nodes while minimizing changes elsewhere in the graph. This selective approach allows the attacker to achieve their objectives with minimal disruption to the overall graph topology. Conversely, non-targeted attacks, exemplified by methods like Mettack~\cite{Zugner2019Mettack}, are aimed at degrading the overall performance of the GNN model across the entire graph. In these attacks, the adversary seeks to compromise the model's ability to make accurate predictions for a wide range of nodes, rather than focusing on specific targets. Whether through targeted or non-targeted means, adversaries seek to exploit vulnerabilities in GNN architectures to manipulate their behavior and compromise their performance.

Adversarial attacks on GNNs vary in their impact and complexity, with some forms of manipulation posing greater challenges than others. In fact, altering the graph structure tends to have a more pronounced effect on GNN performance compared to manipulating node features alone~\cite{Wu2019GCNJaccard}. Moreover, attacks conducted during the training phase (i.e., poisoning attacks) are often more detrimental and difficult to mitigate than those occurring during testing (i.e., evasion attacks)~\cite{Zhu2021relationship}. Given these observations~\cite{Alchihabi2023ELRGNN}, our main focus in this work is on defending against poisoning attacks that target the graph structure, albeit we also consider features attacks for the sake of completeness. Poisoning attacks during training present major challenges due to their potential to compromise the learning process and adversely impact the model's generalization performance. By manipulating the graph structure, adversaries can introduce subtle yet impactful perturbations that persist throughout the training phase, leading to compromised model behavior and reduced predictive accuracy. By prioritizing defense against poisoning attacks, which represent some of the most challenging threats to GNN integrity, our work aims to develop a robust model that can effectively mitigate the impact of adversarial manipulation on the graph structure.

\medskip\noindent\textbf{Defense Strategies for GNNs.} In response to adversarial attacks, considerable efforts have recently been dedicated to devising defensive mechanisms that bolster the robustness of GNNs. For instance, Robust GCN (RGCN)~\cite{Zhu2019RGCN} is a defense method that utilizes Gaussian distributions to represent node embeddings and incorporates a variance-based attention mechanism. Similar to the graph attention (GAT) model~\cite{Velivckovic2018GAT} that extends the fundamental aggregation function of GCN by assigning varying importance to each edge using attention coefficients, the attention mechanism of RGCN assigns weights to node neighborhoods based on their variances. Larger variances indicate a higher likelihood of being targeted by attacks. By leveraging these variances, RGCN penalizes the attention scores of adversarial edges, thereby reducing the propagation of adversarial effects through the graph. GNN-Jaccard~\cite{Wu2019GCNJaccard} is a pre-processing defensive mechanism designed to enhance model robustness against adversarial attacks by removing edges from the graph that exhibit low Jaccard similarity based on the assumption that the clean graph adheres to homophily, where nodes with similar attributes are more likely to be connected.  GCN-SVD~\cite{Entezari2020GCNSVD} is another defense mechanism that operates at the preprocessing stage, specifically targeting high-rank perturbations like those induced by Nettack~\cite{Zugner2018Nettack}. It operates by performing a low-rank approximation of the graph adjacency matrix through truncating its singular values via SVD. By focusing on retaining the low-frequency structural features while eliminating high-frequency perturbations, GCN-SVD aims to enhance the robustness of GNNs against adversarial attacks. Both GNN-Jaccard and GCN-SVD adopt a two-stage preprocessing approach to mitigate the effects of adversarial attacks. In the first stage, these methods preprocess the perturbed graphs to extract clean graph representations. This preprocessing step involves applying specific strategies to identify and remove perturbations introduced by adversarial attacks, such as edges with low Jaccard similarity in GNN-Jaccard or high-frequency components in the adjacency matrix in GCN-SVD. Once the clean graph representations are obtained, the second stage involves training the GCN model on these clean graphs.

As an effective approach, several defense methods propose to leverage the properties of low rank, sparsity, and feature smoothness in the graph structure~\cite{Jin2020ProGNN,Alchihabi2023ELRGNN}. For instance, Pro-GNN~\cite{Jin2020ProGNN} employs a joint learning framework to simultaneously learn the clean graph structure from perturbed data and optimize the parameters of the GNN. It imposes constraints on the graph structure, enforcing it to be low-rank and sparse through regularization, aligning it closely with the clean structure. However, the joint optimization process in Pro-GNN requires iterative updates to both the adjacency matrix and the model parameters, leading to increased computational complexity. GNNGuard~\cite{Zhang2020GNNGUARD} employs a defense strategy against adversarial attacks by assigning higher weights to edges connecting similar nodes and lower weights to edges connecting dissimilar nodes. The rationale behind this approach lies in the assumption that similar nodes are more likely to interact with each other. Although GNNGuard offers promising strategies for defending against adversarial attacks, it relies on the estimation of neighbor importance for every node and the memory layer for graph coarsening, which can incur significant computational overhead, especially for large graphs.Zhu \emph{et al.}~\cite{Zhu2024RNCGLN} address joint robustness to graph and label noise using a graph contrastive objective (local learning), self-attention (global learning), and pseudo graphs/labels for denoising. Their pipeline performs training-time denoising and self-training. In contrast, Fix-GCN is a lightweight architectural defense; robustness comes from its fixed-point-derived higher-order propagation and flexible-pass spectral profile, without graph reconstruction or pseudo labels. Wu \emph{et al.}~\cite{Wu2024GLSGNN} propose a robust anti-fraud framework with an attack simulator and low-rank subspace learning via SVD with task-specific joint losses. Their method is domain-driven (financial fraud) and emphasizes low-rank modeling. Fix-GCN is task-agnostic and avoids matrix factorization while matching GCN complexity.

While incorporating defense mechanisms into GNNs has demonstrated efficacy in countering adversarial attacks, a recent and promising line of work involves the design and development of robust GNNs by devising novel graph filters or feature aggregation schemes tailored to preserve information robustness against adversarial manipulations~\cite{Jin2021SimpGCN,Liu2021ElasticGCN,Chang2021LPF,Lei2022EvenNet,Huang2023MidGCN}. For instance, Simp-GCN~\cite{Jin2021SimpGCN} employs an adaptive message aggregation mechanism to integrate the graph structure and node features, as well as self-supervised learning to capture intricate relationships between node features, including both similarities and dissimilarities. Mid-GCN~\cite{Huang2023MidGCN} employs a mid-pass graph filter to protect against adversarial attacks by leveraging the observation that the eigenvalues of the graph Laplacian within mid-frequency are less affected by such attacks, as mid-frequency signals tend to preserve information from higher-order neighbors. Unlike previous methods, our proposed Fix-GNN model derives its message-passing mechanism from solving a graph filtering system via fixed-point iteration, selectively attenuating high-frequency components while maintaining computational efficiency. By leveraging a higher-order filter, our model captures information from multi-hop neighbors of graph nodes. Moreover, Fix-GCN  incorporates by design a residual connection that ensures information from the initial feature matrix is preserved.

\section{Method} \label{Method}

\subsection{Preliminaries and Problem Formulation}
An attributed graph is a type of graph data structure where each node in the graph is associated with attributes or features. Let $\mathcal{G}=(\mathcal{V},\mathcal{E},\bm{X})$ be an attributed graph, where $\mathcal{V}=\{1,\ldots,N\}$ is the set of $N$ nodes and $\mathcal{E}\subseteq \mathcal{V}\times\mathcal{V}$ is the set of edges, and $\bm{X}=(\bm{x}_{1},...,\bm{x}_{N})^{\T}$ an $N\times F$ feature matrix of node attributes (i.e., $\bm{x}_{i}$ is an $F$-dimensional row vector for node $i$). We denote by $\bm{A}$ an $N\times N$ adjacency matrix whose $(i,j)$-th entry is equal to 1 if $i$ and $j$ are neighboring nodes, and 0 otherwise. We also denote by $\bm{L}=\bm{I}-\hat{\bm{A}}$ the normalized Laplacian matrix, where $\hat{\bm{A}}=\bm{D}^{-1/2}\bm{A}\bm{D}^{-1/2}$ is the normalized adjacency matrix, $\bm{D}=\op{diag}(\bm{A}\bm{1})$ is the diagonal degree matrix, and $\bm{1}$ is an $N$-dimensional vector of all ones. Since the normalized Laplacian matrix is symmetric positive semi-definite, it admits an eigendecomposition given by $\bm{L}=\bm{U}\bg{\Lambda}\bm{U}^{\T}$, where $\bm{U}=(\bm{u}_1,\dots,\bm{u}_N)$ is an orthonormal matrix whose columns constitute an orthonormal basis of eigenvectors and $\bg{\Lambda}=\op{diag}(\lambda_1,\dots,\lambda_N)$ is a diagonal matrix comprised of the corresponding eigenvalues such that $0=\lambda_1\le\dots\le\lambda_N\le 2$ in increasing order~\cite{Chung1997Spectral}.

\medskip\noindent\textbf{Problem Statement.}\quad Semi-supervised learning in a graph involves predicting the labels of nodes that are not labeled, based on the labels of a small subset of nodes. Specifically, let $\mathcal{V}_{l}\subset\mathcal{V}$ be the set of $N_l$ labeled nodes in $\mathcal{V}$ with associated ground-truth labels in the label set $\mathcal{Y}_{l}=\{\bm{y}_{1},\dots,\bm{y}_{N_{l}}\}$, where $\bm{y}_{i}\in\{0,1\}^{C}$ is the one-hot encoding vector of node $i$ and $C$ is the total number of classes. Let $\mathcal{V}_{u}\subset\mathcal{V}\setminus\mathcal{V}_{l}$ be the set of $N_u$ unlabeled nodes, where $N_l+N_u=N$ and $N_{l}\ll N_{u}$. The goal of semi-supervised node classification is to learn the parameters $\bg{\theta}$ of a graph representation learning (GRL) prediction model $f_{\bg{\theta}}: \mathcal{V}_l \to \mathcal{Y}_l$. This is usually done by minimizing the categorical cross-entropy loss function over the set of labeled nodes
\begin{equation}	
\min_{\bg{\theta}}\sum_{i\in\mathcal{V}_{l}}\mathcal{C}(\bm{y}_{i}, \hat{\bm{y}}_{i})=-\sum_{i\in\mathcal{V}_{l}}\sum_{c=1}^{C}y_{ic}\log(\hat{y}_{ic}),
\end{equation}
where $\bm{y}_{i}\in\mathcal{Y}_{l}$ is the one-hot encoding vector of node $i$, $\hat{\bm{y}}_i=f_{\bg{\theta}}(i)$ is the vector of predicted probabilities of node $i$, and $\mathcal{C}(\bm{y}_i,\hat{\bm{y}}_i)$ is the categorical cross-entropy loss for the $i$th node. Here, $y_{ic}$ is the indicator that the $i$th node belongs to the $c$th class, while $\hat{y}_{ic}$ is the predicted probability that the model associates the $i$th node with class $c$.

\smallskip\noindent Adversarial attacks on GRL models can be carried out through various means, including perturbing the graph structure, node features, or a combination of both. To undermine the performance of a GRL model, an adversarial attacker manipulates the edges and/or node features in the original graph $\mathcal{G}$, resulting in perturbed graphs $\mathcal{G}'=(\mathcal{V},\mathcal{E}',\bm{X})$ or $\mathcal{G}'=(\mathcal{V},\mathcal{E}',\bm{X}')$, where $\bm{A}'$ denotes the adjacency matrix of the perturbed graph. In the context of these attacks, $\bm{X}$ represents the original node features, and $\bm{X}'$ represents the manipulated node features. The attacker has the flexibility to target the graph structure (edges) and/or the node features to create perturbations that can deceive the GRL model into making incorrect predictions. Hence, it is crucial to develop a robust GRL model to counter such attacks. Then, the learned robust model is employed to predict the labels of the nodes in the set $\mathcal{V}_{u}$.

\subsection{Spectral Modulation Filtering}
\begin{figure}[!htb]
\centering
\includegraphics[width=3.35in]{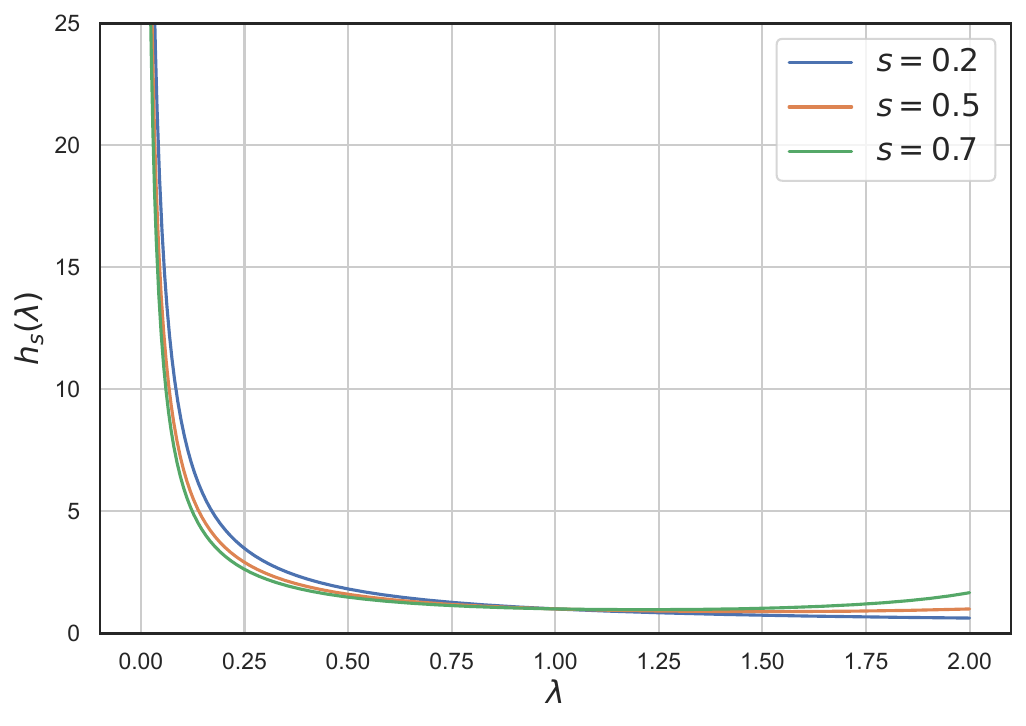}
\caption{Transfer function of the spectral modulation filter. Lower values of the scaling parameter make the filter attenuate high-frequency components more strongly.}
\label{Fig:SHA}
\end{figure}

Spectral graph filtering employs filters defined as functions of the graph normalized Laplacian (or its eigenvalues). The goal of these filters, often referred to as frequency responses or transfer functions, is to reduce or eliminate high-frequency noise in the graph signal. These functions basically describe how a filter affects the input graph signal to produce the output graph signal. We define a spectral modulation filter as follows:
\begin{equation}
h_{s}(\lambda)=\frac{1}{(1+s)\lambda - s\lambda^2},
\end{equation}
where $s\in (0,1)$ is a positive scaling parameter that allows for the modulation or adjustment of the spectral characteristics, indicating its capability to control the filtering effect on different frequency components of the graph signal. The filter $h_s$ is a rational polynomial function of the eigenvalues
of the normalized Laplacian matrix. It is a flexible-pass filter in the sense that it exhibits low-pass characteristics, as it allows low-frequency components (corresponding to small eigenvalues) to flexibly pass through with little attenuation, while attenuFigure2aating high-frequency components (associated with large eigenvalues). As shown in Figure~\ref{Fig:SHA}, the attenuation behavior of the filter is determined by the scalar $s$, which serves as a tuning or modulation parameter. By adjusting $s$, we can control the trade-off between preserving low-frequency structural information and reducing high-frequency noise or variations in the graph signal. When $s$ is large, the filter is less selective, allowing a wider range of frequencies to pass through with less attenuation. As $s$ decreases, the filter becomes more selective and significantly attenuates higher frequencies, effectively filtering out more of the high-frequency noise or variations in the graph signal.

\medskip\noindent\textbf{Graph Filtering System.}\quad Applying the spectral modulation filter on the graph signal $\bm{X}\in\mathbb{R}^{N\times F}$ means filtering each feature channel along the graph spectrum. Formally, the filtered graph signal $\bm{H}$is  given by
\begin{equation}
\bm{H}=h_{s}(\bm{L})\bm{X}=((1+s)\bm{L}-s\bm{L}^{2})^{-1}\bm{X},
\end{equation}
where $h_s(\lambda)=\frac{1}{(1+s)\lambda - s\lambda^2}$ is a rational spectral response and $\mathbf{L}=\mathbf{I}-\hat{\mathbf{A}}$ is the normalized Laplacian. Left-multiplying both sides by $((1+s)\bm{L}-s\bm{L}^2)$ yields the linear system
\begin{equation}
((1+s)\bm{L}-s\bm{L}^{2})\bm{H}=\bm{X},
\end{equation}
or equivalently
\begin{equation}
\begin{split}
\bm{H} &=(\bm{I}-(1+s)\bm{L}+s\bm{L}^{2})\bm{H}+\bm{X} \\
       &=(\bm{I}-s\bm{L})(\bm{I}-\bm{L})\bm{H}+\bm{X}.
\end{split}
\end{equation}
Since $\bm{L}=\bm{I}-\hat{\bm{A}}$, the spectral modulation filter equation becomes
\begin{equation}
\bm{H} = \underbrace{\big((1-s)\bm{I}+s\hat{\bm{A}}\big)\hat{\bm{A}}}_{\bm{P}}\bm{H}+\bm{X},
\label{Eq:SHA}
\end{equation}
which can be solved using, for instance, the fixed point iteration method~\cite{Burden2015Numerical}, where $\bm{H}=\varphi(\bm{H})$ with the function $\varphi$  defined by the right-hand side term of Eq. \eqref{Eq:SHA}. The propagation matrix $\bm{P}$ combines 1-hop and 2-hop diffusion on $\hat{\bm{A}}$ and reveals a fixed-point form amenable to iterative solution.

\medskip\noindent\textbf{Fixed Point Iterative Solution.}\quad Fixed-point iteration is an iterative numerical method used to find a fixed point of a given function~\cite{Burden2015Numerical}. The process involves repeatedly applying the function to an initial guess or estimate and updating this estimate in each iteration until it converges to the fixed point. For the spectral modulation filter equation $\bm{H}=\varphi(\bm{H})$, The fixed point iterative solution is given by
\begin{equation}
\bm{H}^{(t+1)}=((1-s)\bm{I}+s\hat{\bm{A}})\hat{\bm{A}}\bm{H}^{(t)}+\bm{X},
\label{Eq:FP}
\end{equation}
with some initial guess $\bm{H}^{(0)}$, and $t\in\mathbb{N}$ denotes the iteration number.

\subsection{Fixed-Point Iterative Graph Neural Network}
At the core of graph neural networks is the concept of feature propagation rule, which determines how information is passed between nodes in a graph. It involves updating the current node features by aggregating information from their immediate and high-order neighboring nodes, followed by a non-linear activation function to produce an updated representation for the node. Inspired by the fixed point iterative solution of the spectral modulation filter equation, we propose a fixed-point graph convolutional network (Fix-GCN) with the following layer-wise update rule for node feature propagation:
\begin{equation}
\bm{H}^{(\ell+1)}=\sigma\Bigl(((1-s)\bm{I}+s\hat{\bm{A}})\hat{\bm{A}}\bm{H}^{(\ell)}\bm{W}^{(\ell)}
+\bm{X}\widetilde{\bm{W}}^{(\ell)}\Bigr),
\label{Eq:IS}
\end{equation}
where $\bm{W}^{(\ell)}$ and $\widetilde{\bm{W}}^{(\ell)}$ are learnable weight matrices, $\sigma(\cdot)$ is an element-wise activation function, $\bm{H}^{(\ell)}\in\mathbb{R}^{N\times F_{\ell}}$ is the input feature matrix of the $\ell$-th layer with $F_{\ell}$ feature maps for $\ell=0,\dots,L-1$. The input of the first layer is the initial feature matrix $\bm{H}^{(0)}=\bm{X}$. It is worth pointing out that the key difference between Eq.~\eqref{Eq:FP} and Eq.~\eqref{Eq:IS} is that the latter defines a representation updating rule for propagating node features layer-wise using trainable weight matrices for learning an efficient representation of the graph, followed by an activation function to introduce non-linearity into the network in a bid to enhance its expressive power.

The update rule of Fix-GCN is essentially comprised of three main components: (i) feature propagation that combines the features of the 1- and 2-hop neighbors of nodes (i.e., it aggregates information from immediate and high-order neighboring nodes), (ii) feature transformation that applies learnable weight matrices to the node representations to learn an efficient representation of the graph, and (iii) residual connection for ensuring that information from the initial feature matrix is preserved. The initial residual connection used in the proposed model allows information from the initial feature matrix to bypass the current layer and be directly added to the output of the current layer. This helps preserve important information that may be lost during the aggregation process, thereby improving the flow of information through the network. In other words, in addition to performing a second-order graph convolution, the update rule of Fix-GCN applies an initial residual connection that reuses the initial node features. Note that the propagation operation or matrix $\bm{P}=((1-s)\bm{I}+s\hat{\bm{A}})\hat{\bm{A}}$ of the proposed GNN is a weighted combination of the normalized adjacency matrix and its square. It allows Fix-GCN to capture information from nodes that are not only directly connected (1-hop), but also incorporates information from the neighbors of the neighbors (2-hop). The parameter $s$ helps control the balance between the information from immediate neighbors and the information from nodes that are at most two edges away in the graph. This is particularly valuable for learning graph representations that capture more global information and dependencies.

\medskip\noindent\textbf{Connection to GCN.}\quad The hyper-parameter $s$ in Fix-GCN provides control over the network's behavior. When $s=0$, Fix-GCN reduces to the standard GCN with initial residual connections, which basically decouples the transformations for the self-connections and the 1-hop neighbors, essentially incorporating residual connections at the initial stage of the graph convolution process. This process operates on the normalized adjacency matrix without self-connections. Similarly, when $s=1$, Fix-GCN corresponds to the second-order GCN with initial residual connections. So, by varying the value of $s$ between 0 and 1, we can smoothly control the behavior of Fix-GCN, allowing it to capture different orders of information from the graph structure. Therefore, Fix-GCN provides a flexible framework for adapting to various graph-based tasks and the level of emphasis on local (first-order) and non-local (second-order) information in the graph structure.

\medskip\noindent\textbf{Model Complexity.} \quad For simplicity, we assume the feature dimensions are the same across all layers, i.e., $F_{\ell}=F$ for all $\ell$, with $F \ll N$. Multiplying the propagation matrix $((1-s)\bm{I}+s\hat{\bm{A}})\hat{\bm{A}}$ with an embedding $\bm{H}^{(\ell)}$ costs $\mathcal{O}(\Vert\hat{\bm{A}}\Vert_{0}F)$ in time, where $\Vert\hat{\bm{A}}\Vert_{0}$ denotes the number of non-zero entries of the sparse matrix $\hat{\bm{A}}$ (i.e., number of edges in the graph). Multiplying an embedding with a weight matrix costs $\mathcal{O}(NF^2)$. Also, multiplying the initial feature matrix by the residual connection weight matrix costs $\mathcal{O}(NF^2)$. Hence, the time complexity of an $L$-layer Fix-GCN is $\mathcal{O}(L\Vert\hat{\bm{A}}\Vert_{0}F+LNF^2)$.

For memory complexity, an $L$-layer Fix-GCN requires $\mathcal{O}(LNF+LF^2)$ in memory, where $\mathcal{O}(LNF)$ is for storing all embeddings and $\mathcal{O}(LF^2)$ is for storing all layer-wise weight matrices. Therefore, our proposed Fix-GCN model has the same time and memory complexity as that of GCN, albeit Fix-GCN takes into account both immediate and distant graph nodes for improved learned node representations. It is important to note that there is no need to explicitly compute the square of the normalized adjacency matrix in the Fix-GCN model. Instead, we perform right-to-left multiplication of the normalized adjacency matrix with the embedding. This process avoids the computational cost associated with matrix exponentiation and simplifies the computation, making our model more efficient while achieving its objectives. For sparse graphs with bounded average degree, Fix-GCN scales linearly in $N$ for fixed $L,F$, matching standard GCN while providing higher-order, fixed-point-derived propagation.

\medskip\noindent\textbf{Numerical Stability.} \quad To demonstrate the numerical stability of the proposed Fix-GCN model, we start with a useful result in matrix analysis~\cite{Riesz1990FA}, which states that the spectral radius of the sum of two commuting matrices is bounded by the sum of the spectral radii of the individual matrices.

\begin{lemma}
If two matrices $\bm{C}_{1}$ and $\bm{C}_{2}$ commute, i.e., $\bm{C}_{1}\bm{C}_{2}=\bm{C}_{2}\bm{C}_{1}$, then
$$\rho(\bm{C}_{1}+\bm{C}_{2})\le \rho(\bm{C}_{1}) + \rho(\bm{C}_{2}),$$
where $\rho(\cdot)$ denotes matrix spectral radius (i.e., largest absolute value of all eigenvalues).
\end{lemma}

Since the eigenvalues of the normalized Laplacian matrix $\bm{L}=\bm{I}-\hat{\bm{A}}$ lie in the interval $[0,2]$, it follows that $\rho(\hat{\bm{A}})\le 1$. Hence, we have the following result, which demonstrates the training stability of the proposed model, with information smoothly propagating through the graph layers without amplifying or dampening effects that could lead to instability.

\begin{proposition}
The update rule of Fix-GCN is numerically stable.
\end{proposition}
\noindent\textit{Proof.}\quad Recall that the propagation matrix of Fix-GCN is given by $$\bm{P}=((1-s)\bm{I}+s\hat{\bm{A}})\hat{\bm{A}}=(1-s)\hat{\bm{A}}+s\hat{\bm{A}}^2.$$
Since the matrices $(1-s)\hat{\bm{A}}$ and $s\hat{\bm{A}}^2$ satisfy the assumptions of Lemma 1, we have
$$\rho((1-s)\hat{\bm{A}}+s\hat{\bm{A}}^2)\le \rho((1-s)\hat{\bm{A}}) + \rho(s\hat{\bm{A}}^2)\le 1,$$
because both $\rho(\hat{\bm{A}})$ and $\rho(\hat{\bm{A}}^2)$ are bounded by 1. Hence, the spectral radius of the propagation matrix is bounded by 1. Consequently, repeated layer-wise application of this propagation operator is stable, meaning that it will not cause the node feature representations to diverge. This stability is crucial for maintaining control over the feature representations as they propagate through the network, ensuring that the model remains numerically well-behaved and learns meaningful patterns from the graph data.

\section{Experiments} \label{Experiments}
In this section, we conduct experimental evaluations of the proposed model, comparing it with state-of-the-art methods.

\subsection{Experimental Setup}
\noindent\textbf{Datasets.}\quad We assess the performance of our proposed method on five benchmark datasets: GitHub~\cite{rozemberczki2021multi}, Cora-ML~\cite{bojchevski2017deep}, and citation networks (Cora, CiteSeer, PubMed)~\cite{sen2008collective}. Dataset statistics are summarized in Table~\ref{Tab:dataset}, where only the largest connected component is considered.

\begin{table}[!htb]
\caption{Summary statistics of benchmark graph datasets. We only consider the largest connected component in these adversarial graphs.}
\medskip
\centering
\begin{tabular}{l l l l c}
\toprule
Datasets & \#Nodes & \#Edges & \#Features & \#Classes\\
\midrule
Cora & 2,485 & 5,069 & 1,433 & 7\\
CiteSeer & 2,110 & 3,668 & 3,703 & 6 \\
Cora-ML & 2,995 & 4,208 & 2,879 & 5 \\
GitHub & 3,150 & 71,310 & 4,005 & 2 \\
PubMed & 19,717 & 44,338 & 500 & 3 \\
\bottomrule
\end{tabular}
\label{Tab:dataset}
\end{table}

\medskip\noindent\textbf{Baseline Methods.}\quad We evaluate the performance of our model against comparative GNN models and state-of-the-art adversarial defense methods, including GCN~\cite{Kipf2017GCN}, GAT~\cite{Velivckovic2018GAT}, GCN-Jaccard~\cite{Wu2019GCNJaccard}, GCN-SVD~\cite{Entezari2020GCNSVD}, RGCN~\cite{Zhu2019RGCN}, Pro-GNN~\cite{Jin2020ProGNN}, and Mid-GCN~\cite{Huang2023MidGCN}.

\medskip\noindent\textbf{Implementation Details.}\quad All experiments are conducted on a Linux machine with a single NVIDIA GeForce RTX 3070 GPU featuring 8GB of memory. For fair comparison, we run experiments following the setup and default settings of the baselines, including the data split for semi-supervised learning. For each dataset, we randomly assign 10\% of the nodes to the labeled training set, 10\% of the nodes to the validation set, and the remaining 80\% of the nodes to the test set. We use PyTorch to implement our two-layer model with a hidden dimension of 64, and train it for 200 epochs using the Adam optimizer~\cite{Kingma2015Adam} with a learning rate of 1e-2 and a weight decay rate of 5e-4. The default dropout ratio is 0.6 for all datasets, and the filter hyperparameter $s = 0.2$ is determined via grid search.

\subsection{Results and Analysis}
We evaluate the node classification performance of Fix-GCN against various types of adversarial attacks, including
non-targeted attacks, targeted attacks, random attacks, and feature attacks.

\medskip\noindent\textbf{Robustness Against Non-targeted Attacks.}\quad The goal of non-targeted attacks is to degrade the overall performance of the mode. We adopt Mettack~\cite{Zugner2019Mettack} as a non-targeted attack to perturb the graph structure with the aim of compromising the model's performance in node classification. Specifically, we evaluate the robustness of our model against non-targeted adversarial attacks using different perturbation rates, spanning from 0\% to 25\% in increments of 5\%. The node classification results of Fix-GCN and baseline methods are summarized in Table~\ref{mettack_table}, where both the average accuracy and standard deviation are reported over 10 runs. The best results are in bold and the second best ones are underlined. The results presented in Table~\ref{mettack_table} show that, for the vast majority of cases, our model consistently surpasses all baselines across all datasets, with notable performance improvements observed, especially at higher perturbation rates. At 25\% perturbation rate, our Fix-GCN model yields relative improvements of 6.4\%, 1.98\%, 2.97\% and 11.73\% over Mid-GCN on Cora, CiteSeer, GitHub and PubMed, respectively, with the highest relative improvement of 13.23\% achieved on Cora-ML. Interestingly, our model outperforms Mid-GCN under all perturbation rates on the dense GitHub dataset, which has the highest number of edges among the five graph benchmarks.

\begin{table*}[!htb]
\caption{Node classification performance of Fix-GCN and baselines under non-targeted attacks (Mettack) with different perturbation rates P(\%). We report the average accuracy over 10 runs, along with the corresponding standard deviation. The best results are in \textbf{bold} and the second best ones are \underline{underlined}.}
\medskip
\centering
\small 
\setlength\tabcolsep{6pt} 
\begin{tabular}{c|c|cccccccc}
\toprule
Dataset & P(\%) & GCN & GAT & GCN-Jaccard & GCN-SVD & RGCN & Pro-GNN & Mid-GCN & \textbf{Fix-GCN}\\
\midrule
\multirow{5}{*}{\rotatebox[origin=c]{90}{Cora}}
& 0 & 83.50$\pm$0.44 & 83.97$\pm$0.69 & 82.05$\pm$0.51 & 80.63$\pm$0.45 & 83.09$\pm$0.44 & \textbf{85.39$\pm$0.81} & 84.61$\pm$0.46 & \underline{84.80$\pm$0.33} \\
& 5 & 76.55$\pm$0.79 & 80.44$\pm$0.74 & 79.13$\pm$0.59 & 78.93$\pm$0.53 & 77.42$\pm$0.39 & 82.78$\pm$0.39 & \textbf{82.94$\pm$0.46} & \underline{82.19$\pm$0.27} \\
& 10 & 70.39$\pm$1.28 & 75.61$\pm$0.59 & 75.16$\pm$0.76 & 71.47$\pm$0.83 & 72.22$\pm$0.38 & 79.03$\pm$0.59 & \underline{80.14$\pm$0.86} & \textbf{82.64$\pm$0.58} \\
& 15 & 65.10$\pm$0.71 & 69.78$\pm$1.28 & 71.03$\pm$0.64 & 66.69$\pm$1.18 & 66.82$\pm$0.39 & 76.40$\pm$1.27 & \underline{77.77$\pm$0.75} & \textbf{80.58$\pm$0.81} \\
& 20 & 59.56$\pm$0.92 & 59.54$\pm$0.92 & 65.71$\pm$0.89 & 58.94$\pm$1.13 & 59.27$\pm$0.37 & 73.32$\pm$1.56 & \underline{76.58$\pm$0.29}  & \textbf{79.27$\pm$0.55} \\
& 25 & 47.53$\pm$1.96 & 54.78$\pm$0.74 & 60.82$\pm$1.08 & 52.06$\pm$1.19 & 50.51$\pm$0.78 & 69.72$\pm$1.69 & \underline{72.89$\pm$0.81} & \textbf{77.56$\pm$0.94} \\

\midrule
\multirow{5}{*}{\rotatebox[origin=c]{90}{CiteSeer}}
& 0 & 71.96$\pm$0.55 & 73.26$\pm$0.83 & 72.10$\pm$0.63 & 70.65$\pm$0.32 & 71.20$\pm$0.83 & 73.28$\pm$0.69 &  \textbf{74.17$\pm$0.28} & \underline{73.68$\pm$0.31}
\\ & 5 & 70.88$\pm$0.62 & 72.89$\pm$0.83 & 70.51$\pm$0.97 & 68.84$\pm$0.72 & 70.50$\pm$0.43 & 73.09$\pm$0.34 & \textbf{74.31$\pm$0.42} & \underline{74.03$\pm$0.22}
\\ & 10 & 67.55$\pm$0.89 & 70.63$\pm$0.48 & 69.54$\pm$0.56 & 68.87$\pm$0.62 & 67.71$\pm$0.30 & 72.51$\pm$0.75 & \underline{73.59$\pm$0.29} & \textbf{74.27$\pm$0.31}
\\ & 15 & 64.52$\pm$1.11 & 69.02$\pm$1.09 & 65.95$\pm$0.94 & 63.26$\pm$0.96 & 65.69$\pm$0.37 & 72.03$\pm$1.11 & \underline{73.69$\pm$0.29}& \textbf{73.82$\pm$1.02}
\\ & 20 & 62.03$\pm$3.49 & 61.04$\pm$1.52 & 59.30$\pm$1.40 & 58.55$\pm$1.09 & 62.49$\pm$1.22 & 70.02$\pm$2.28 & \underline{71.51$\pm$0.83} & \textbf{72.80$\pm$0.47} \\
& 25 & 56.94$\pm$2.09 & 61.85$\pm$1.12 & 59.80$\pm$1.47 & 57.18$\pm$1.87 & 55.35$\pm$0.66 & 68.95$\pm$2.78 & \underline{69.12$\pm$0.72} & \textbf{70.49$\pm$0.69} \\

\midrule
\multirow{5}{*}{\rotatebox[origin=c]{90}{Cora-ML}}
& 0 & 85.85$\pm$0.30 & 83.45$\pm$1.46 & 80.22$\pm$1.54 & 83.87$\pm$1.53 & 82.39$\pm$1.71 & 85.38$\pm$1.14 & \underline{86.56$\pm$0.28} & \textbf{86.78$\pm$0.56}\\
& 5 & 79.76$\pm$1.44 & 79.11$\pm$1.69 & 79.75$\pm$1.78 & 80.29$\pm$1.89 & 80.13$\pm$1.91 & \textbf{80.38$\pm$1.98} & \underline{80.20$\pm$1.68} & 80.07$\pm$0.66 \\
& 10 & 74.64$\pm$0.67 & 79.36$\pm$1.66 & 74.33$\pm$1.79 & 79.08$\pm$1.73 & 74.55$\pm$1.91 & \textbf{79.48$\pm$1.23} & \underline{79.30$\pm$0.96} & 79.17$\pm$0.82\\
& 15 & 53.74$\pm$1.09 & 61.22$\pm$1.44 & 57.38$\pm$1.09 & 74.95$\pm$1.58 & 55.42$\pm$1.25 & 53.60$\pm$1.18 & \underline{73.32$\pm$0.66} & \textbf{76.95$\pm$1.42}\\
& 20 & 45.24$\pm$1.88 & 52.72$\pm$1.29 & 47.15$\pm$1.09 & 47.95$\pm$1.76 & 47.68$\pm$1.32 & 47.37$\pm$1.34 & \underline{60.92$\pm$1.43} & \textbf{75.17$\pm$1.24} \\
& 25 & 48.80$\pm$1.91 & 54.26$\pm$1.98 & 49.42$\pm$1.56 & 56.85$\pm$1.99 & 50.62$\pm$1.01 & 50.52$\pm$1.12 & \underline{67.18$\pm$1.35} & \textbf{76.09$\pm$0.67} \\

\midrule
\multirow{5}{*}{\rotatebox[origin=c]{90}{GitHub}}
& 0 & 72.92$\pm$0.13 & 72.81$\pm$0.12 & 72.93$\pm$0.56 & 73.31$\pm$0.15 & 73.16$\pm$0.19 & 73.34$\pm$0.34 &  \underline{79.51$\pm$0.67} & \textbf{81.39$\pm$0.23} \\
& 5 & 72.81$\pm$0.07 & 72.43$\pm$1.13 & 71.85$\pm$2.18 & 72.91$\pm$0.12 & 73.09$\pm$0.31 & 72.89$\pm$0.07 & \underline{81.87$\pm$1.46} & \textbf{82.59$\pm$0.42}
\\ & 10 & 72.61$\pm$0.57 & 72.97$\pm$0.13 & 72.63$\pm$0.98 & 72.78$\pm$0.09 & 73.06$\pm$0.16 & 72.75$\pm$0.18 & \underline{81.23$\pm$1.67} & \textbf{82.45$\pm$0.35}
\\ & 15 & 72.97$\pm$0.11 & 72.97$\pm$0.06 & 72.79$\pm$0.51 & 72.97$\pm$1.13 & 73.22$\pm$0.21 & 72.98$\pm$0.09 & \underline{80.48$\pm$0.25} & \textbf{82.75$\pm$0.26}
\\ & 20 & 72.11$\pm$2.27 & 70.42$\pm$2.12 & 72.24$\pm$1.96 & 72.47$\pm$0.14 & 73.10$\pm$0.21 & 72.98$\pm$0.13 & \underline{81.08$\pm$0.96} & \textbf{82.62$\pm$0.28} \\
& 25 & 72.74$\pm$0.41 & 72.97$\pm$1.16 & 72.32$\pm$1.73 & 72.14$\pm$0.07 & 72.91$\pm$0.24 & 72.56$\pm$0.21 & \underline{80.37$\pm$1.51} & \textbf{82.76$\pm$0.30} \\

\midrule
\multirow{5}{*}{\rotatebox[origin=c]{90}{PubMed}}
& 0 & 87.19$\pm$0.09 & 83.73$\pm$0.40 & 87.06$\pm$0.09 & 83.44$\pm$0.21 & 86.16$\pm$0.18 & \underline{87.33$\pm$0.18} & 85.67$\pm$0.37 & \textbf{88.26$\pm$0.27}\\
& 5 & 83.09$\pm$0.13 & 78.00$\pm$0.44 & 86.39$\pm$0.06 & 83.41$\pm$0.15 & 81.08$\pm$0.20 & \underline{87.25$\pm$0.09} & 83.48$\pm$0.10 & \textbf{87.33$\pm$0.38}\\
& 10 & 81.21$\pm$0.09 & 74.93$\pm$0.38 & 85.70$\pm$0.07 & 83.27$\pm$0.21 & 77.51$\pm$0.27 & \textbf{87.25$\pm$0.09} & 81.43$\pm$0.43 & \underline{86.77$\pm$0.25}\\
& 15 & 78.66$\pm$0.12 & 71.13$\pm$0.51 & 84.76$\pm$0.08 & 83.10$\pm$0.18 & 73.91$\pm$0.25 & \textbf{87.20$\pm$0.09} &  79.74$\pm$0.14 & \underline{86.33$\pm$0.12}\\
& 20 & 77.35$\pm$0.19 & 68.21$\pm$0.96 & 83.01$\pm$0.22 & 83.88$\pm$0.05 & 71.18$\pm$0.31 & \underline{87.15$\pm$0.15} & 78.69$\pm$0.32 & \textbf{87.28$\pm$0.21} \\
& 25 & 75.50$\pm$0.17 & 65.41$\pm$0.77 & 83.66$\pm$0.06 & 82.72$\pm$0.18 & 67.95$\pm$0.15 & \underline{86.76$\pm$0.19} & 77.81$\pm$0.34 & \textbf{86.94$\pm$0.16} \\
\bottomrule
\end{tabular}
\label{mettack_table}
\end{table*}

\begin{figure}[!htb]
\hspace*{.3cm}\includegraphics[width=3.4in]{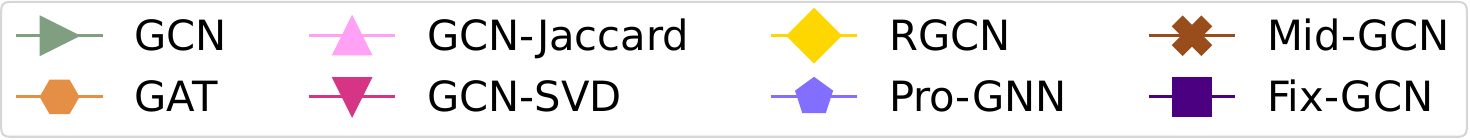}
\centering
\setlength{\tabcolsep}{1.5pt}
\begin{tabular}{cc}
\includegraphics[width=1.73in]{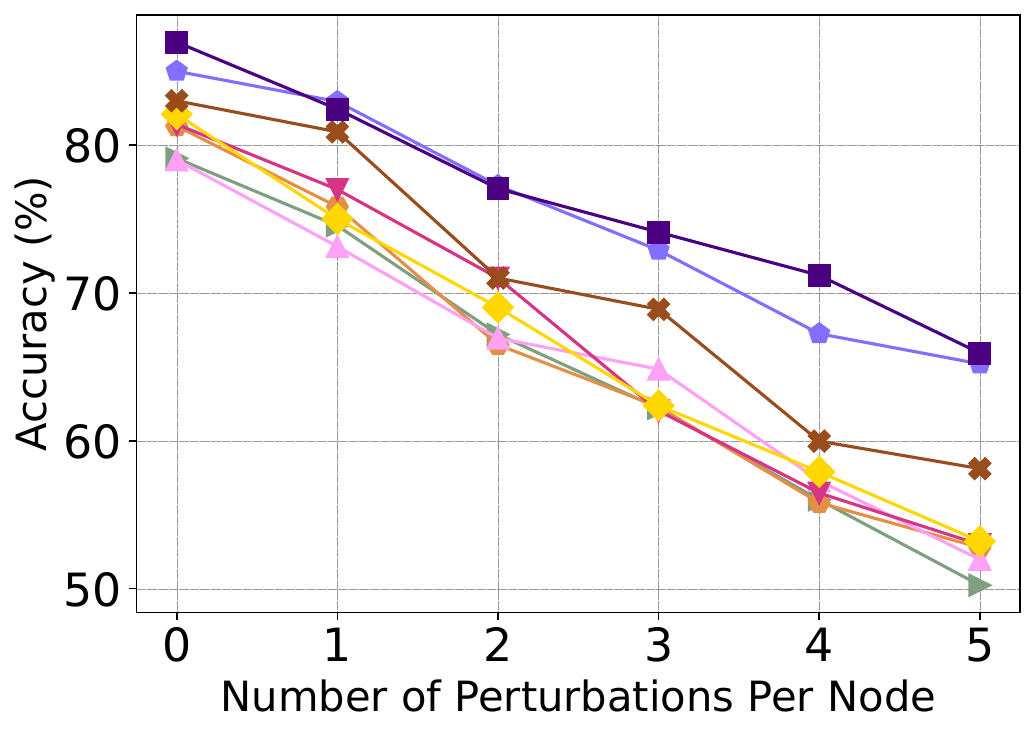} & \includegraphics[width=1.73in]{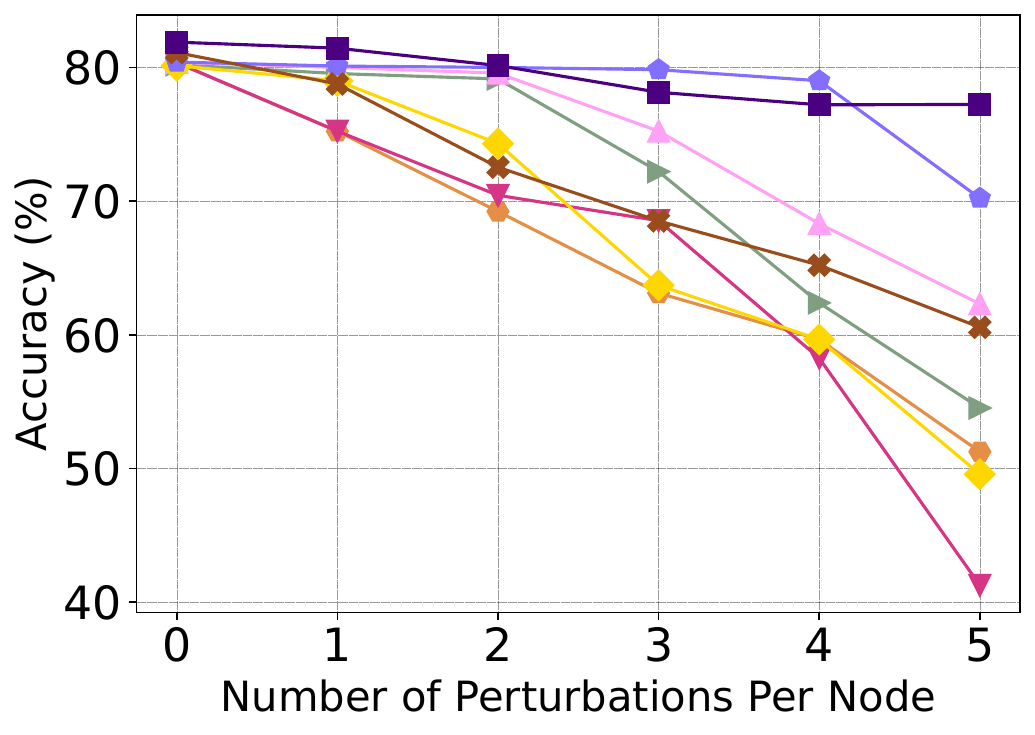} \\
(a) Cora & (b) CiteSeer  \\
\includegraphics[width=1.73in]{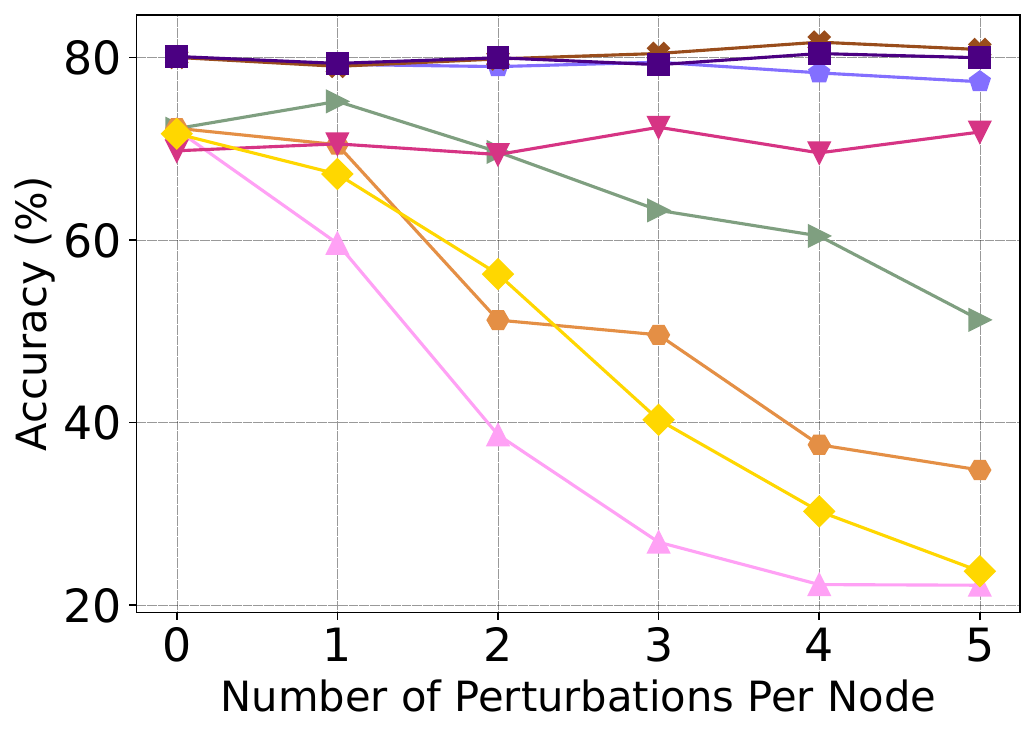} & \includegraphics[width=1.73in]{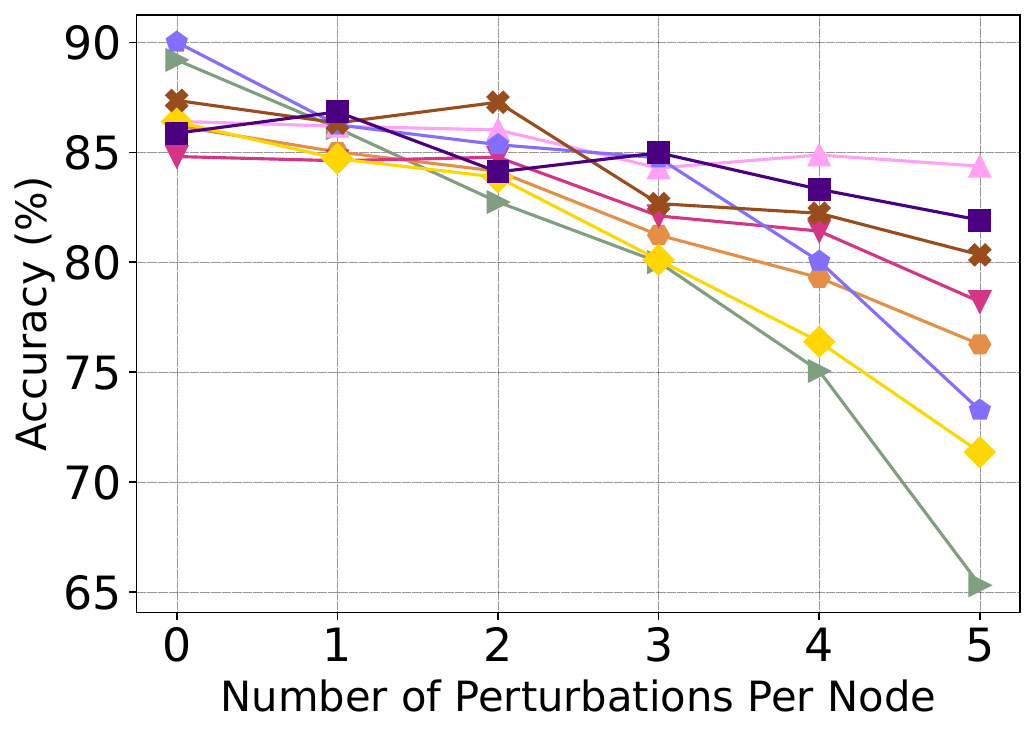} \\
(c) GitHub & (d) PubMed
\end{tabular}
\caption{Node classification accuracy under targeted attacks (Nettack) with varying numbers of perturbations on the target nodes $\{1, 2, 3, 4, 5\}$.}
\label{Nettack}
\end{figure}

\medskip\noindent\textbf{Robustness Against Targeted Adversarial Attacks.}\quad Unlike non-targeted attacks, which degrade the overall performance of the model, targeted attacks aim to cause the model to produce incorrect predictions for specific nodes. For instance, a targeted attack might involve perturbing the graph structure in a way that causes the model to misclassify a particular node as belonging to a specific class, even if it does not naturally belong to that class. We adopt Nettack\cite{Zugner2018Nettack} as a targeted adversarial attack method, which iteratively perturb the graph structure by removing or adding edges in a way that maximally changes the predictions of the GNN model for the target nodes. For this attack, consistent with prior work~\cite{Zhu2019RGCN,Jin2020ProGNN}, we change the number of perturbations made on each targeted node from 1 to 5, incrementally increasing by 1. Nodes in the test set with a degree exceeding 10 are designated as target nodes. Since the GitHub dataset exhibits a relatively high level of density, we opt to select only 8\% of the nodes for attacks. Similarly, for the PubMed dataset, we only use 10\% of the nodes in our analysis to prevent potential lengthy execution times associated with Nettack. The defense performance results (i.e., multi-class classification accuracy) against targeted attacks are depicted in Figure~\ref{Nettack}, which shows that our model consistently outperforms the baseline methods under the varying numbers of perturbations on the target nodes across all the datasets. The main findings from Figure~\ref{Nettack} are summarized as follows:

\begin{itemize}
\item When compared to the strongest baselines, Mid-GCN and Pro-GNN, our Fix-GCN model consistently achieves superior performance on the Cora dataset across all levels of perturbations. Similarly, on the CiteSeer dataset, Fix-GCN outperforms Mid-GCN in most cases, especially as the number of perturbations increases. Moreover, it is noteworthy that GCN-SVD, despite being tailored for Nettack, exhibits a significant drop in performance, particularly at higher levels of perturbations.
\item On the GitHub dataset, known for its denser nature, our model, alongside the strongest baselines, Mid-GCN and Pro-GNN, demonstrates consistent performance across all levels of perturbations. Also, GCN-SVD maintains a stable performance due to its specific design for Nettack. A similar trend is observed on the PubMed dataset, albeit with a slight decline in performance, particularly at higher levels of perturbations. Interestingly, GCN-Jaccard demonstrates superior performance at higher levels of perturbations, showcasing the effectiveness of its two-stage approach. However, it is important to note that this approach involves preprocessing the input graph by dropping dissimilar edges based on the Jaccard similarity metric before training GCN on the processed graph. Notably, without this preprocessing step, the GCN baseline experiences a significant drop in performance as the number of perturbations increases.
\item By combining the strengths of a robust aggregation mechanism that captures information from higher-order neighbors of graph nodes and a flexible-pass filtering approach that preserves low-frequency structural information in the graph signal, our Fix-GCN model outperforms all the baselines in most cases.
\end{itemize}
These results underscore the robustness and effectiveness of our Fix-GCN model in defending against adversarial attacks across various settings and datasets.

\begin{figure}[!htb]
\hspace*{.3cm}\includegraphics[width=3.4in]{Plots_Legend.pdf}
\centering
\setlength{\tabcolsep}{1.5pt}
\begin{tabular}{cc}
\includegraphics[width=1.73in]{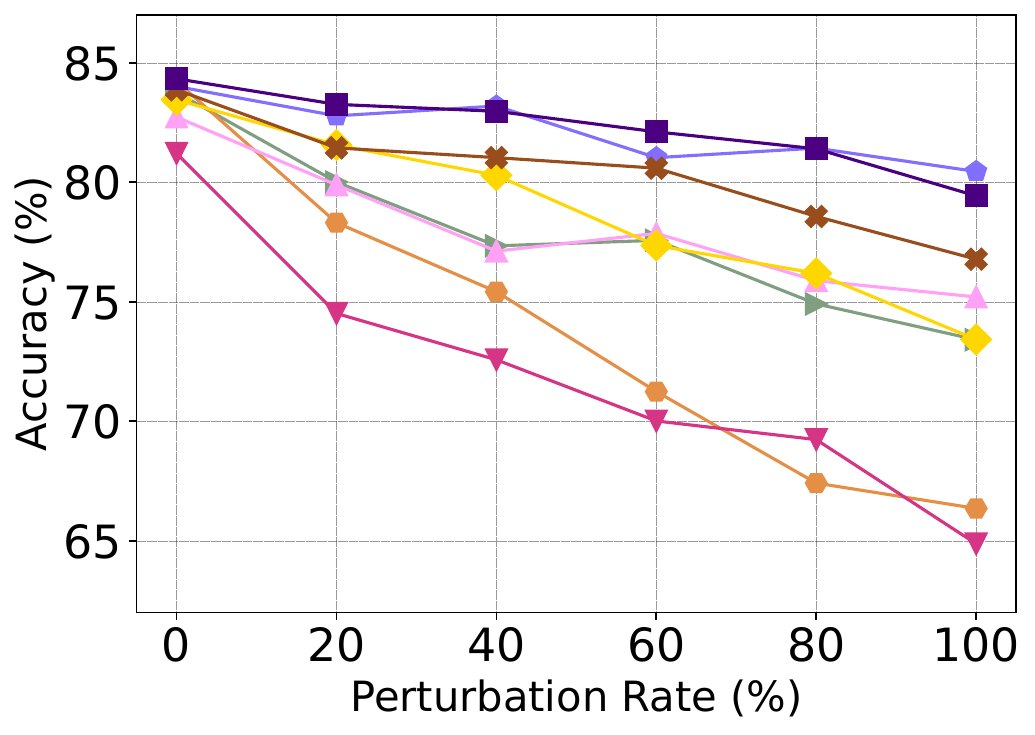} & \includegraphics[width=1.73in]{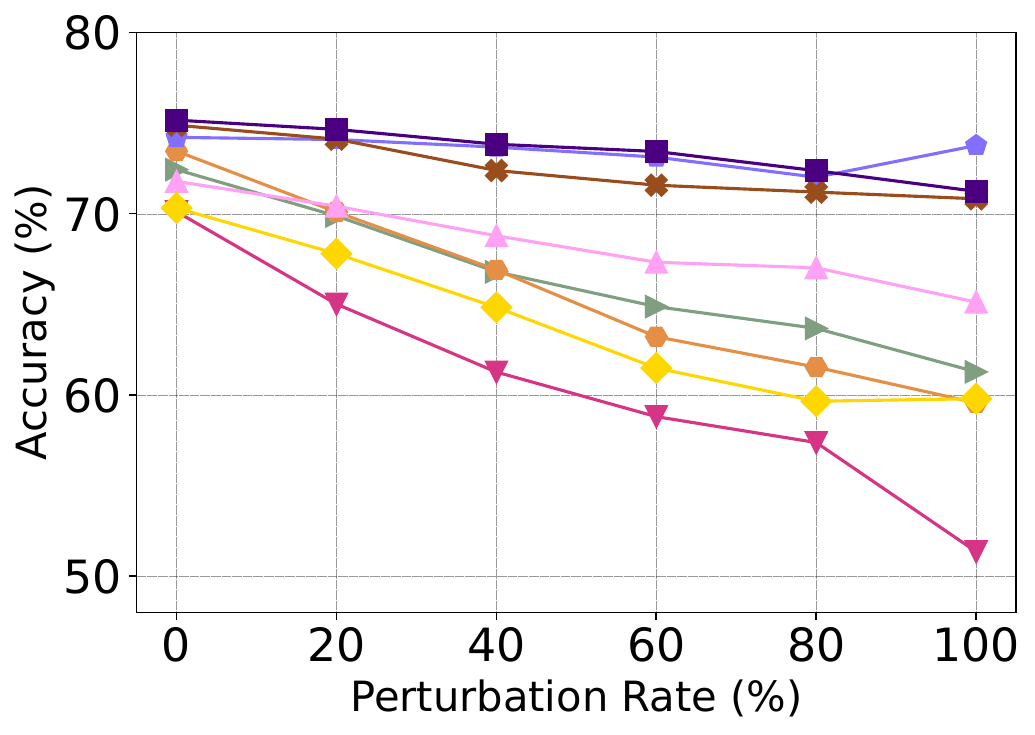} \\
(a) Cora & (b) CiteSeer
\end{tabular}
\caption{Node classification accuracy under random attacks with varying perturbation rates.}
\label{Random}
\end{figure}

\medskip\noindent\textbf{Robustness Against Random Attacks.}\quad The objective of random attacks is to introduce randomness by adding edges to the input graph, akin to injecting random noise into the clean graph. In our experiment, we assess the performance of our Fix-GCN model and baseline methods under varying ratios of random attack, ranging from 0\% to 100\% of the number of edges in the true adjacency matrix, with increments of 20\%. The results, reported in Figure~\ref{Random}, show that Fix-GCN, along with Mid-GCN and Pro-GNN, maintains relatively stable performance across all perturbation rates on Cora and CiteSeer datasets. Notably, Fix-GCN exhibits superior results on most perturbation rates for both datasets, except at the highest 100\% rate. In contrast, the other comparative methods experience significant performance degradation as the perturbation rate increases.

\begin{figure}[!htb]
\hspace*{.3cm}\includegraphics[width=3.4in]{Plots_Legend.pdf}
\centering
\setlength{\tabcolsep}{1.5pt}
\begin{tabular}{cc}
\includegraphics[width=1.73in]{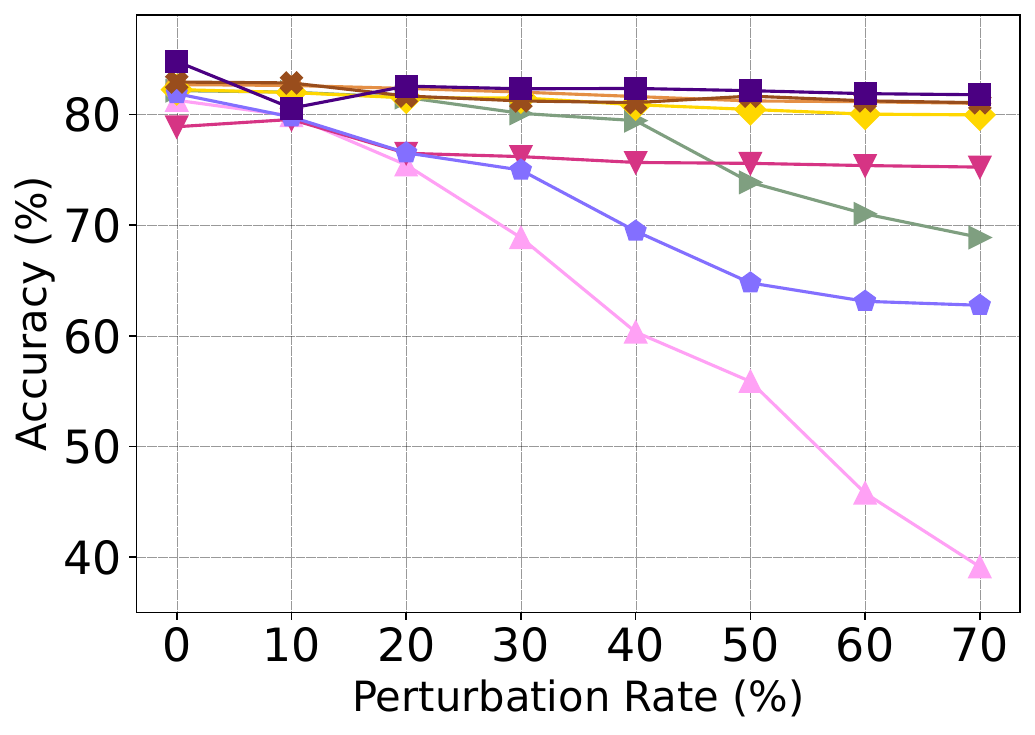} & \includegraphics[width=1.73in]{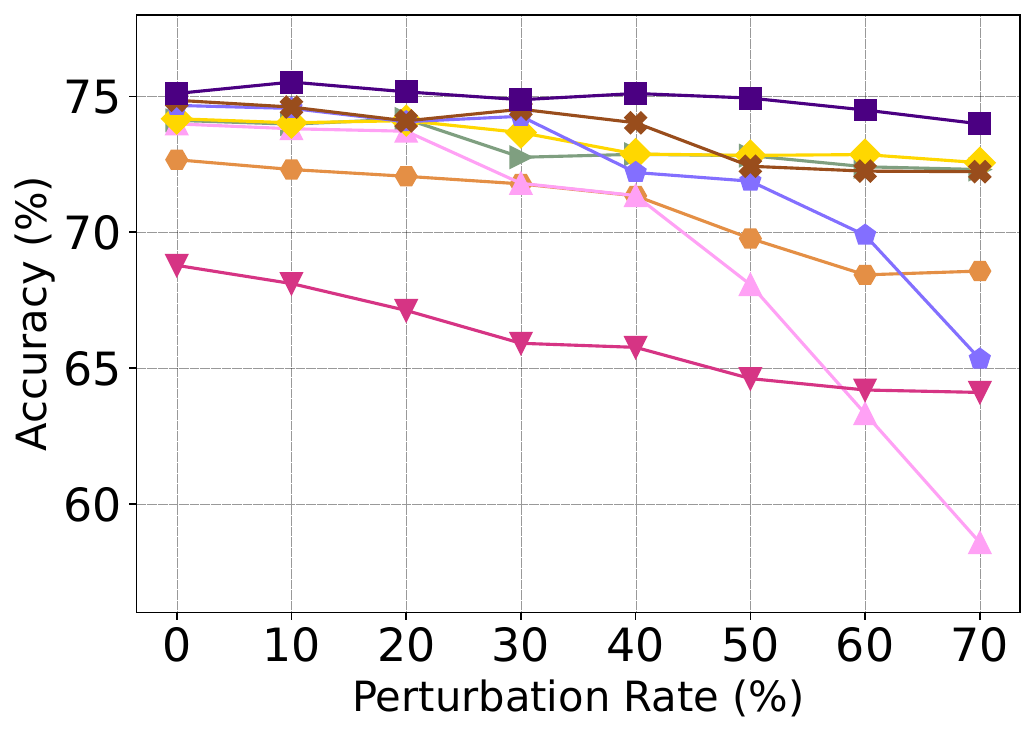} \\
(a) Cora & (b) CiteSeer
\end{tabular}
\caption{Node classification accuracy under feature attacks with different perturbation rates.}
\label{Feature_Attack}
\end{figure}

\medskip\noindent\textbf{Robustness Against Feature Attacks.}\quad We assess the influence of random attacks on the efficacy of our Fix-GCN model by randomly perturbing the initial node feature matrix. Besides structural alterations to graphs, feature attacks represent a crucial aspect of adversarial attacks since node features are extensively leveraged by GCN-based methods in their message-passing mechanisms. Given that the features of Cora, CiteSeer, and GitHub datasets are exclusively comprised of 0s and 1s, we introduce feature attacks by randomly flipping the 0/1 values~\cite{zhang2020feature}. For instance, the Cora and CiteSeer graphs are composed of nodes representing scientific publications and edges representing citation links between these publications. Each node is described by a binary feature vector indicating the presence or absence of words from a dictionary. The results, as illustrated in Figure~\ref{Feature_Attack}, demonstrate that Fix-GCN consistently outperforms all baseline methods across all perturbation rates, particularly at higher rates, on both datasets. Interestingly, both Mid-GCN and Pro-GNN experience significant performance drops at higher levels of perturbation rates. A more pronounced decline in performance is observed for GCN-Jaccard on both datasets. In contrast, our Fix-GCN model demonstrates robust defense against feature attacks. This strong resilience against such attacks is largely attributed to the fact that Fix-GCN integrates an initial residual connection in its feature propagation scheme by design. The initial residual connection in the proposed model serves the crucial function of preserving information from the initial feature matrix throughout the aggregation process. By allowing the initial features to directly contribute to the output of each layer, the residual connection ensures that important information is retained, even in the face of adversarial perturbations targeting the node features. In essence, the initial residual connection acts as a safeguard against feature manipulation, enhancing the model's resilience to adversarial attacks on node features.

\begin{figure}[!htb]
\hspace*{.3cm}\includegraphics[width=3.4in]{Plots_Legend.pdf}
\centering
\setlength{\tabcolsep}{1.5pt}
\begin{tabular}{cc}
\includegraphics[width=1.73in]{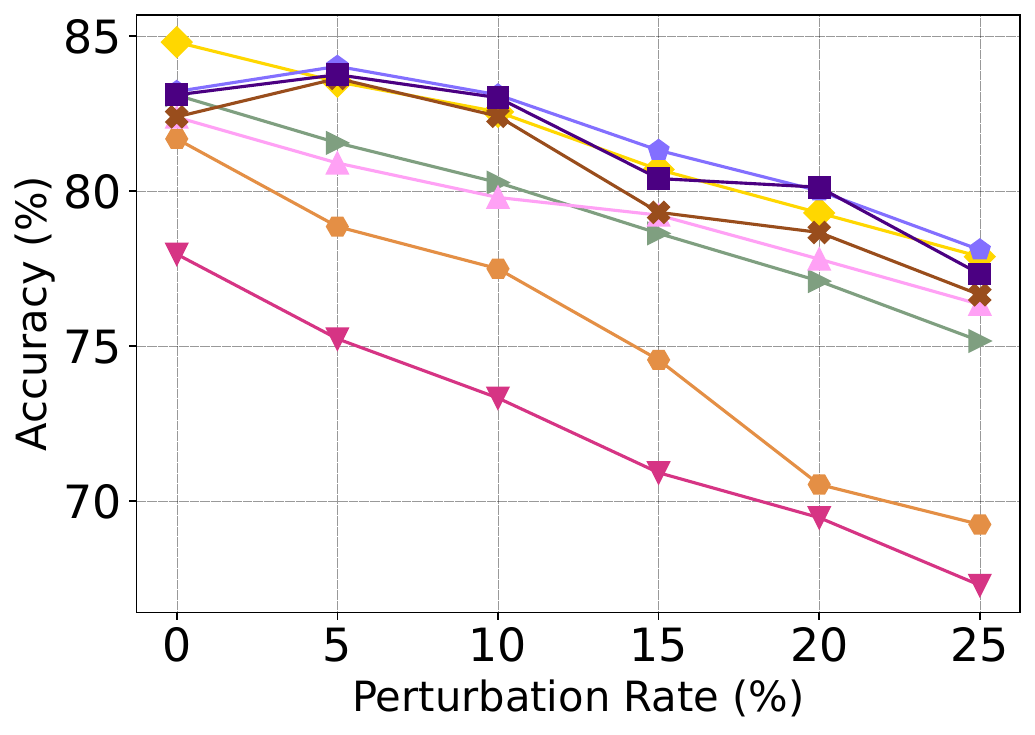} & \includegraphics[width=1.73in]{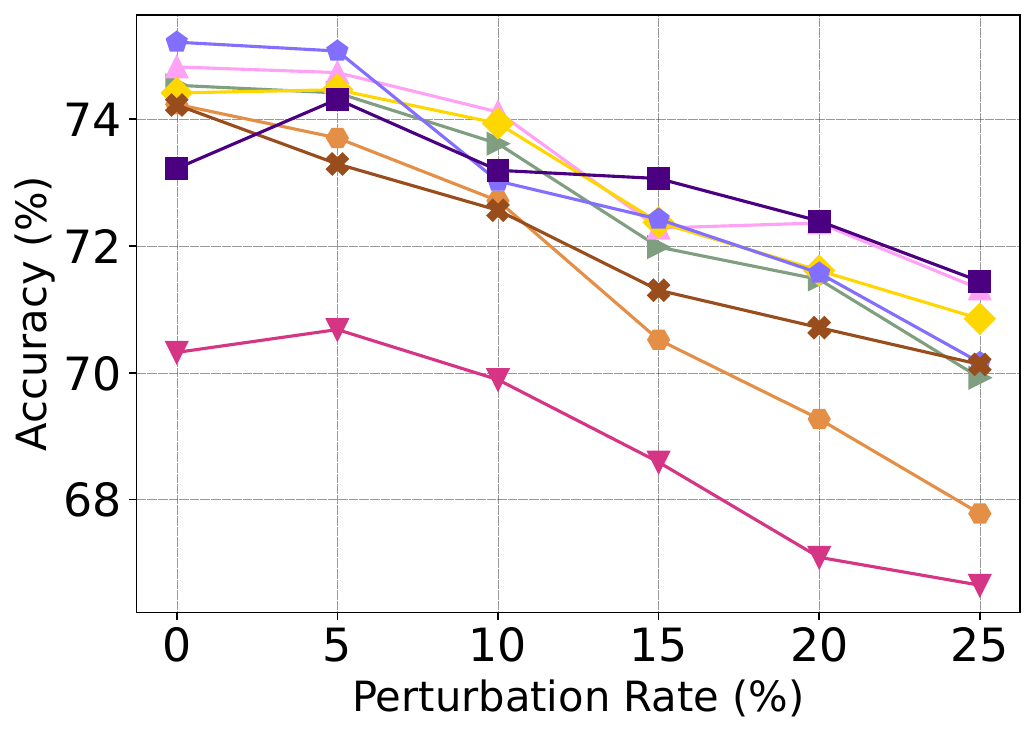} \\
(a) Cora & (b) CiteSeer
\end{tabular}
\caption{Node classification accuracy of different models under evasion attacks (DICE) with varying perturbation rates.}
\label{DICE}
\end{figure}

\medskip\noindent\textbf{Robustness Against Evasion Attacks.}\quad The objective of evasion attacks is to manipulate the model predictions by making small, often imperceptible changes to the graph structure during the testing phase.  To assess the vulnerability of our model to evasion attacks, we employ a variant of the disconnect internally, connect externally (DICE) method~\cite{waniek2018hiding,Lei2022EvenNet}, which is a white box attack strategy. This method involves randomly connecting nodes with different labels or dropping edges between nodes that share the same label. In this experiment, we vary the perturbation rate from 0\% to 25\%, with a step size of 5\%, to evaluate the performance of our Fix-GCN model against evasion attacks. The results depicted in Figure~\ref{DICE} illustrate that Fix-GCN demonstrates robustness against evasion attacks, particularly at higher perturbation rates, where it outperforms Mid-GCN by a significant margin across both datasets. Furthermore, the performance of GAT and GCN-SVD drops rapidly as the perturbation rate increases, highlighting the effectiveness of Fix-GCN in defending against evasion attacks.

\begin{figure}[!htb]
\centering
\includegraphics[width=3.2in]{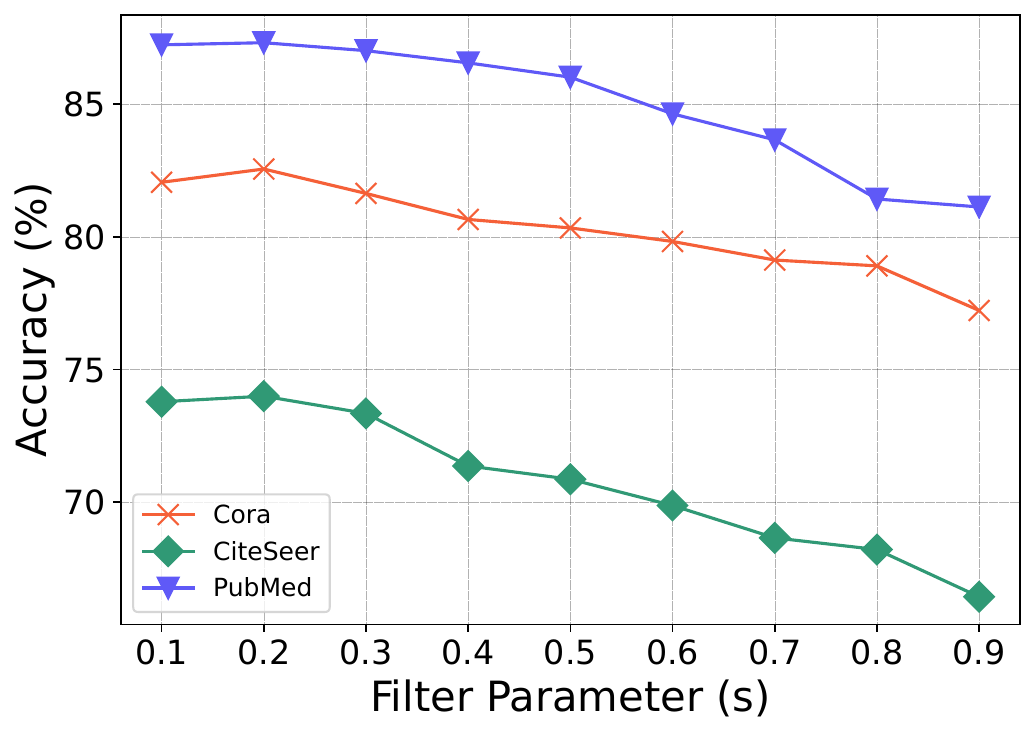}
\caption{Node classification accuracy on citation networks (Cora, CiteSeer, PubMed) under Mettack with a 5\% perturbation rate using various values of the spectral modulation filtering parameter $s$.}
\label{Parameter}
\end{figure}

\medskip\noindent\textbf{Parameter Sensitivity Analysis.}\quad We study the performance variation for our model on the three citation networks with respect to the spectral modulation filtering parameter $s$. We vary $s$ from 0.1 to 0.9, and the results are presented in Figure~\ref{Parameter} using Mettack as an adversarial attack with a 5\% perturbation rate. It is evident that the accuracy remains relatively stable when $s$ fluctuates between 0.1 and 0.3, which correspond to low-pass filtering. The best performance is achieved when $s=0.2$, which is the value that we set in our experiments.

\section{Discussion}
In this section, we outline the merits of the proposed Fix-GCN model in three key aspects:
\begin{itemize}
\item \textit{Flexibility.} By leveraging a flexible-pass filter that selectively attenuates high-frequency components while preserving low-frequency structural information in the graph signal, our Fix-GCN model is able to effectively mitigate the impact of adversarial perturbations. As $s$ increases toward 1, the transfer function becomes less selective, allowing more mid/high-frequency content to pass. In addition, Fix-GCN's initial residual path injects the raw features at each layer, which helps retain sharper (potentially high-frequency) signals useful for heterophily-like patterns. Moreover, capturing information from higher-order neighbors reduces the susceptibility of our model to direct perturbations on 1-hop neighbors of target nodes. Adversarial attacks that directly manipulate the immediate neighbors of target nodes can significantly impact the model's performance. By incorporating information from higher-order neighbors, Fix-GCN can mitigate the effects of such direct perturbations. On the other hand, if the adversarial perturbations primarily corrupt low/mid bands, we can learn $s$ per layer or learn a small set of coefficients that linearly combine $\hat{\bm{A}}$ and $\hat{\bm{A}}^2$, yielding a data-adaptive spectral profile while maintaining Fix-GCN's efficiency.
\item \textit{Robustness.} Fix-GCN's combination of selective filtering, higher-order information capture, initial residual connection, and stable performance makes it robust against various forms of adversarial attacks. In particular, the ability of our model to capture information from higher-order neighbors adds an additional layer of defense against adversarial attacks, enhancing the robustness of Fix-GCN in real-world scenarios where targeted perturbations on immediate neighbors may occur frequently.
\item \textit{Efficiency.} The time and memory complexity of Fix-GCN is on the same order as that of the standard GCN despite considering both immediate and distant graph nodes for improved node representations. This computational efficiency is achieved without the need for explicitly computing the square of the normalized adjacency matrix, as Fix-GCN utilizes right-to-left matrix multiplication to compute its second-order propagation matrix. This ensures that our model maintains practicality and scalability for real-world applications, as it can handle large-scale datasets efficiently without significantly increasing computational overhead.
\end{itemize}
\textbf{Limitations:}\quad While our model demonstrates robust performance against various adversarial attacks, particularly at higher perturbation levels, it is worth pointing out its three main limitations. First, the proposed flexible-pass filter attenuates high frequencies more as $s$ decreases. On graphs/tasks where high-frequency components are crucial (e.g., heterophily), a purely low-pass bias can be suboptimal. Adaptively weighting high-frequency components can help on heterophilous graphs. Second, while our sensitivity study shows a stable plateau for $s\in[0.1,0.3]$, the exact best value is not universal across datasets. Third, while we avoid explicitly forming \(\hat{\mathbf{A}}^2\), each layer performs two sparse multiplies with $\hat{\bm{A}}$. For very dense graphs (growing average degree), per-epoch time increases accordingly.

\section{Conclusion} \label{Conclusion}
In this paper, we introduced Fix-GCN, a robust model against adversarial attacks. The core concept behind Fix-GCN lies in its message-passing mechanism, which is derived from solving a spectral graph modulation filtering system using fixed-point iteration. One of the key strengths of our model is its ability to capture information from higher-order connections, thereby improving its resilience against direct perturbations on immediate neighbors, which are often more vulnerable to adversarial attacks. By considering information from higher-order neighbors, Fix-GCN effectively captures the global context of the graph, making it more resilient to direct perturbations on 1-hop neighbors. Moreover, Fix-GCN maintains computational efficiency comparable to the standard GCN without sacrificing performance or requiring additional defense mechanisms. Our comparative experiments showed that our model yields improved performance compared to strong baselines across different graph datasets and under a variety of adversarial attacks. As future work, we intend to apply the proposed model to other downstream tasks such as anomaly detection.

\section*{Compliance with Ethical Standards}
\noindent{\textbf{Conflict of Interest}}\quad The authors declare that they have no financial or personal interests to disclose.

\smallskip\noindent{\textbf{Funding}}\quad This work was supported in part by Natural Sciences and Engineering Research Council of Canada.

\smallskip\noindent{\textbf{Data Availability}}\quad The datasets used in the experiments are publicly available.

\section*{Acknowledgments}
This work was supported in part by the Natural Sciences and Engineering Research Council of Canada (NSERC). Researchers funded through the NSERC-CSE Research Communities Grants do not represent the Communications Security Establishment Canada or the Government of Canada. Any research, opinions or positions they produce as part of this initiative do not represent the official views of the Government of Canada.

\bibliographystyle{ieeetr}
\bibliography{references}
\end{document}